# Towards automated website classification by Deep Learning

*Fabrizio De Fausti, Francesco Pugliese, Diego Zardetto* [1]

## Abstract

*In recent years, the interest in Big Data sources has been steadily growing within the Official Statistics community. The Italian National Institute of Statistics - Istat is currently carrying out several Big Data studies. One of these studies, the ICT Big Data project, resulted in 2018 in the publication of a first set of experimental statistics on the activities that enterprises carry out through their websites (web ordering, job vacancy advertisement, use of social media, etc.). This project aims at exploiting massive amounts of textual data automatically scraped from the websites of Italian enterprises in order to predict a set of target variables (e.g. "e-commerce") that are routinely observed by the traditional ICT Survey. In this paper, we show that Deep Learning techniques can successfully address this problem. Essentially, we tackle a text classification task: an algorithm must learn to infer whether an Italian enterprise performs e-commerce from the textual content of its website. To reach this goal, we developed a sophisticated processing pipeline and evaluated its performance through extensive experiments. Our pipeline uses Convolutional Neural Networks and relies on Word Embeddings to encode raw texts into grayscale images (i.e. normalised numeric matrices). Web-scraped texts are huge and have very low signal to noise ratio: to overcome these issues, we adopted a framework known as False Positive Reduction, which has seldom (if ever) been applied before to text classification tasks. Several original contributions enable our processing pipeline to reach good classification results. Empirical evidence shows that our proposal outperforms all the alternative Machine Learning solutions already tested in Istat for the same task.*

**Keywords:** Deep Learning, Convolutional Neural Networks, Big Data, Web Scraping, Text Classification, Word Embeddings, False Positive Reduction.

---

1  Fabrizio De Fausti (defausti@istat.it); Francesco Pugliese (frpuglie@istat.it); Diego Zardetto (zardetto@istat.it), Italian National Institute of Statistics - Istat.

*The views and opinions expressed are those of the authors and do not necessarily reflect the official policy or position of the Italian National Institute of Statistics - Istat.*

*The authors would like to thank the anonymous reviewers for their comments and suggestions, which enhanced the quality of this article.*

---

## 1. Introduction

In recent years, a new research field known under the name of *Big Data* has emerged (Manyika *et al.*, 2011). In common meaning, Big Data are huge, heterogeneous collections of data sets that are difficult to handle by using state-of-the-art data processing approaches and traditional data management tools (*e.g.* relational databases). Doug Laney popularised "Volume, Velocity and Variety" (known as 3Vs) to characterise the concept of Big Data (Laney, 2001). "Volume" refers to the size of data sets, "Velocity" indicates the speed of data flows, and "Variety" describes the diversity of data types and sources. Big Data continuously grow in size because they are increasingly being generated by disparate sources, such as people posting messages on social networks, users publishing documents on the Web, ubiquitous information-sensing mobile devices, search engines and software logs, sensor networks, e-commerce and stock market transactions, large-scale scientific experiments, and so on. Owing to the speed of these data sets' growth, devising effective methods for processing and analysing Big Data is still an open problem. An even bigger challenge is, perhaps, the one of extracting useful insights from this typically unstructured, noisy and heterogeneous mass of information.

In the last seven years, the Official Statistics community has been converging towards the consensus that Big Data must be one of the pillars of the ongoing modernisation efforts put in place by National Statistical Institutes (NSIs). For instance, the "Scheveningen Memorandum" (ESS, DGINS, 2013) acknowledges that Big Data sources represent new opportunities and challenges for the European Statistical System (ESS), therefore compelling European NSIs to explore the potential of Big Data for the production of official statistics. At present, similar initiatives are in place worldwide within the statistical systems of all advanced countries.

The Italian National Institute of Statistics - Istat is currently carrying out several Big Data studies, as schematically reported in Table 1. In the present paper we will focus on the "*ICT Big Data project*". This project essentially aims at exploiting textual data automatically scraped from the websites of Italian enterprises in order to predict a set of target variables (*e.g.* "e-commerce") falling within the scope of the Italian "Survey on ICT Usage in Enterprises". The motivation of the project lies in the fact that a powerful and reliable prediction model could be applied, after careful validation, to the whole target population of the ICT survey (as of 2014, about 190,000 enterprises according to Istat's Business Register ASIA, 70% of which was estimated to own a website by the 2014 round of the ICT survey). This, in turn, would enable Istat to: (*i*) enrich the Italian Business Register, and (*ii*) increase the quality of the output estimates produced by the ICT survey (*e.g.* by reducing their Mean Square Error via composite estimators). This line of research eventually resulted in late 2018 in the publication of a first set of experimental statistics on the activities that enterprises carry out through their websites (web ordering, job vacancy advertisement, use of social media, etc.), see *e.g.* Barcaroli and Scannapieco 2019, and references therein.

**Table 1 - Main studies on Big Data currently ongoing at Istat**

| Big Data Source | Official Statistics Domain | Maturity Level |
|---|---|---|
| Scanner data | Consumer Prices | Production |
| Internet data (web scraping) | ICT Usage in Enterprises | Experimental Statistics |
| Social media (Twitter) | Social Mood on Economy | Experimental Statistics |
| Open street map | Road Accidents Indicators | Experimental Statistics |
| Mobile phone data (Call Detail Records) | Mobility and Tourism statistics | Proof of Concept |
| Satellite images | Land Cover Statistics and Maps | Proof of Concept |
| Search engine queries (Google Trends) | Labour Force statistics | Research |
| Traffic cameras (online webcams) | Road Traffic and Accidents statistics | Research |

Source: Our processing

In this work, we will describe a novel, experimental approach to address the prediction task of the ICT Big Data project, based on *Deep Learning* techniques.

The rest of the paper is organised as follows: Section 2 provides examples of Big Data Analytics and outlines the challenges Big Data pose to Machine Learning; Section 3 introduces Deep Learning and discusses it in a Big Data perspective; Section 4 offers background information about Istat's ICT Big Data project, with a focus on its prediction task; Section 5 introduces our Deep Learning proposal and sets our model of choice: *Convolutional Neural Networks*; Section 6 studies the feasibility of our approach, and illustrates preliminary results of a first naïve implementation; Section 7 describes in depth the development of our processing pipeline, from design principles to methods and technical details; Section 8 empirically evaluates the performance of our processing pipeline; Section 9, finally, hints at ongoing work and draws some conclusions.

## 2. Big Data Analytics and Machine Learning

In 2011 there were about 2.5 quintillion bytes of data created every day (Hilbert and Lopez, 2011), and this number still keeps increasing rapidly. Today, Big Data applications are involved in many scientific and industrial fields, and *Big Data Analytics* (Wilder-James, 2012) is often perceived as one of the most relevant business battlefields. Many leading companies all over the world are currently investing resources on sophisticated *Machine Learning* (ML) techniques, striving to mine the value hidden inside Big Data, with the goal of improving pricing strategies and advertising campaigns (Chen and Zhang, 2014). Similarly, the ESS is right now exploring the potential of selected Big Data sources to enhance the quality of existing official statistics, to investigate new phenomena and to produce innovative statistical products. The most promising sources identified so far within the ESSnet Big Data project[2] include: (*i*) enterprise websites (for statistics about job vacancies and enterprise characteristics); (*ii*) smart meters (for statistics about energy, census housing and the environment); (*iii*) AIS, *i.e.* shipboard Automatic Identification System data (for statistics about maritime transport); (*iv*) mobile phone data (for statistics about population, mobility and tourism); (*v*) social media (for statistics about consumer confidence, social tension and economic mood); (*vi*) satellite imagery (for statistics about agriculture and the environment). Just like it is happening in the private sector, ML techniques are destined to play a central role in processing Big Data within NSIs. Interestingly, this is something the whole Official Statistics community agrees on, *despite* the development of sound methodologies to extract valid statistical information from Big Data is still fairly embryonic. However, traditional ML approaches have often shown severe limits when faced with the objective of analysing and exploiting the today's jungle of Big Data. For instance, the fact that many promising Big Data sources generate *unstructured* data, *e.g.* natural language texts (Gentzkow *et al.*, 2017), poses a major challenge to Big Data Analytics. Indeed, most ML approaches require input data to be organised according to a "case by variable" data-model, whereas a natural notion of "variable" no longer exists for entirely unstructured data. As a consequence, meaningful features have to be somehow extracted from raw data before analysis. Unfortunately, human intervention and domain knowledge are needed to perform this data-preparation step, known as *feature engineering*. But feature engineering is costly, impairs automation and hinders scalability. From this premise, the necessity arises for the creation and adoption of a "universal framework" able to deal with the inherent diversity, complexity and hugeness of Big Data, possibly in a highly scalable way. In this respect, Deep Learning naturally emerges as a really promising candidate, owing to its remarkable ability to automatically extract from raw data features that prove useful for a wide range of learning tasks (Goodfellow *et al.*, 2016).

---

2   ESSnet Big Data website: https://webgate.ec.europa.eu/fpfis/mwikis/essnetbigdata/index.php/Main_Page

## 3. Deep Learning

Since around 5 years, *Deep Learning* (DL) is emerging as a promise land for Big Data Analytics, in that it seems excellently fit to distill useful insight from Big Data. Deep Learning typically (though not exclusively) relies on Artificial Neural Networks (ANN). In the past, ANNs have attracted great attention both in mathematical statistics (*e.g.* as universal approximation models of arbitrary nonlinear functions) (Hornik *et al.*, 1989) and in economics (*e.g.* for time series forecasting, see the seminal paper (Kuan and White, 1994) and the literature review (Herbrich *et al.*, 1999).

However, the traditional literature on ANNs mostly focussed on shallow multi-layer feedforward architectures characterised by fully connected neurons, while todays DL research explores a much wider range of alternative topologies and connectivity patterns. Deep Artificial Neural Networks are ANNs characterised by *multiple* hidden layers of neurons which are connected according to a *hierarchical* architecture. Figure 1 depicts the transition from old fashioned neural networks (also known as "Vanilla" or "Shallow" NNs) to *Deep* Neural Networks (DNN). DNNs are right now emerging as a scalable, robust and reliable machine learning paradigm, especially after last years' hardware breakthroughs (GPU computing), new model regularisation (Data-augmentation, Dropout, Batch Normalisation) and training methodologies (Stochastic Gradient Descent, Adam, RMSProp). Specifically, the reason of the success of Deep Learning in Big Data Analytics originates from three major advantages that this approach provides:

- **Robust**: no need to design the features (data representations) ahead of time. Features are automatically learned to be optimal to the task at hand. Robustness to natural variations in the data is automatically learned.

- **Versatile**: the same Deep Learning approach can be used for many different applications and data types. For example, an algorithm performing very well in computer vision will likely lead to pretty good results in speech recognition or in text classification (and vice versa).

- **Scalable**: performance improves with more data. Furthermore the method is massively parallelizable. This makes Deep Learning a perfect fit for parallel computing architectures like GPUs.

**Figure 1 - From Vanilla Neural Networks (one hidden layer) to Deep Neural Networks (multiple hidden layers)**

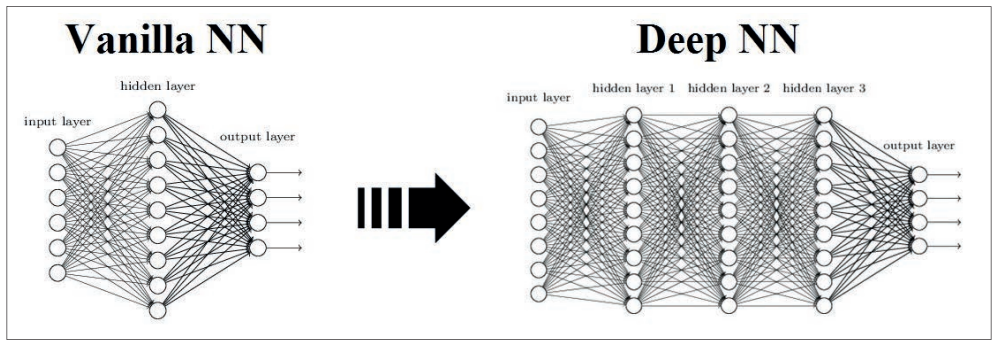

Source: Our processing on Nielsen, 2015

Deep Learning refers to techniques that *automatically* extract meaningful (from a human perspective) complex data representations (features) at high levels of abstraction (Bengio and LeCun, 2007). Such a methodology discovers and learns task-specific data features in a layered, hierarchical fashion. In other words, simpler (less abstract) data features are extracted by lower layers of the neural network and passed to the next layer, where they are combined to form higher-level (more abstract) features, and so on (Bengio *et al.*, 2013). This mechanism is depicted in Figure 2 for a DNN tackling a face recognition task. While the early hidden layers are only able to learn quite simple image features (*e.g.* edges and light or dark spots), these simple features get smartly assembled within the middle hidden layers, where more articulated representations start to form, like eyes, mouths, noses, and other face parts. In the higher hidden layers we can eventually catch sight of raw faces: a "face model" has been distilled, which provides an abstract and high-level synthesis of all the invariant elements of the faces acquired by the DNN during the training phase.

**Figure 2 - Illustration of the features formation within deep hierarchical layers, from less abstract to more abstract data representations**

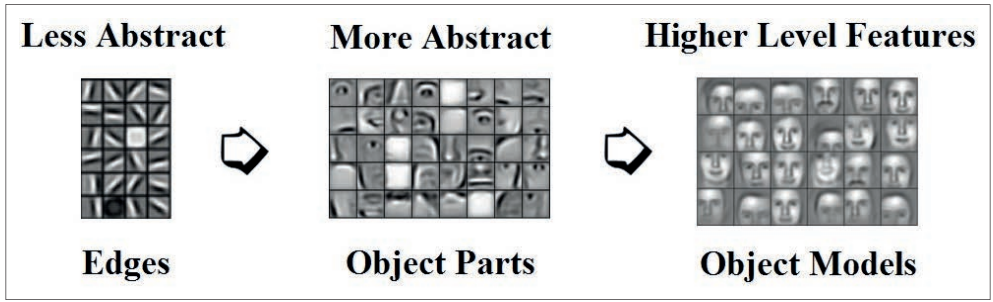

Source: Our processing on Lee *et al.*, 2011

The hierarchical learning architecture of Deep Learning algorithms inspires to the deep layered organisation of the primary sensorial areas of the neocortex in the human brain (which is indeed a *natural* Big Data analyser) (Arel *et al.*, 2010). According to many cognitive scientists, hierarchical processing plays a fundamental role in the cortical computation (Hinton, 2007) and it *must* be the key factor for all the biologically inspired computational models (Riesenhuber, and Poggio, 1999).

From an algorithmic point of view, there is growing empirical evidence that data representations obtained by connecting *many* nonlinear feature extractors in series (as in *Deep* ANNs) generally yields better results as compared to traditional machine learning approaches (Larochelle *et al.*, 2009).

Stated very concisely, the more the layers, the more complicated nonlinear transformations can be learned. Moreover, as shown in Figure 2, DNNs are entirely *data-driven*, namely they extract high-level abstractions and representations *without* any human intervention. On the contrary, classical machine learning algorithms are often unable to identify the complex and nonlinear patterns that are observed in Computer Vision, Speech Recognition and Natural Language Processing tasks, thus requiring feature engineering (*i.e.* human intervention and domain knowledge) to reach effective results.

By automatically extracting features and abstractions from the underlying data, Deep Learning algorithms can address many important problems in Big Data Analytics (Najafabadi *et al.*, 2015). Furthermore, in contrast to more conventional machine learning and feature engineering algorithms, Deep Learning has the advantage of potentially providing a solution to tackle data analysis and learning problems found in massive volumes of *unsupervised* input data (Najafabadi *et al.*, 2015). This makes Deep Learning especially attractive, since nowadays the overwhelming majority of Big Data is made of extremely diverse and complex raw data, which are largely *unlabelled* and *un-categorised* (National Research Council, 2013).

According to the supervised or unsupervised nature of the learning process, DNNs can be divided in two classes:

- **Unsupervised Learning**: In 2006, Hinton proposed deep architectures whose learning process worked in an unsupervised, greedy, layer-wise manner (Hinton *et al.*, 2006). These deep architectures are called *Deep Belief Networks*. Basically, they are a stack of *Restricted Boltzmann Machines* and/or autoassociators called *Autoencoders* (Hinton and Zemel, 1994).

- **Supervised Learning**: In 1989, LeCun proposed the first simple stack of *convolutional* layers and fully connected layers known as LeNet for discriminative and supervised learning purposes (LeCun *et al.*, 1989). This model was enhanced by deeper and more accurate models like *AlexNet* in 2012 (Krizhevsky *et al.*, 2012), *GoogleNet* in 2014 (Szegedy *et al.*, 2015), and Microsoft Research's *Residual Neural Networks* in 2015, which reached super-human capabilities in some Computer Vision tasks (He *et al.*, 2016).

All the supervised learning DL models cited above belong to the *Convolutional Neural Networks* (CNN) family. The connectivity pattern between neurons of CNNs was originally inspired by the organisation of the visual cortex of mammals (see Figure 3). Unsurprisingly, CNNs typically show excellent performance on even very hard image recognition tasks.

**Figure 3 - A schematic representation of a generic Convolutional Neural Network, highlighting the topology and the main building blocks of this kind of deep architecture**

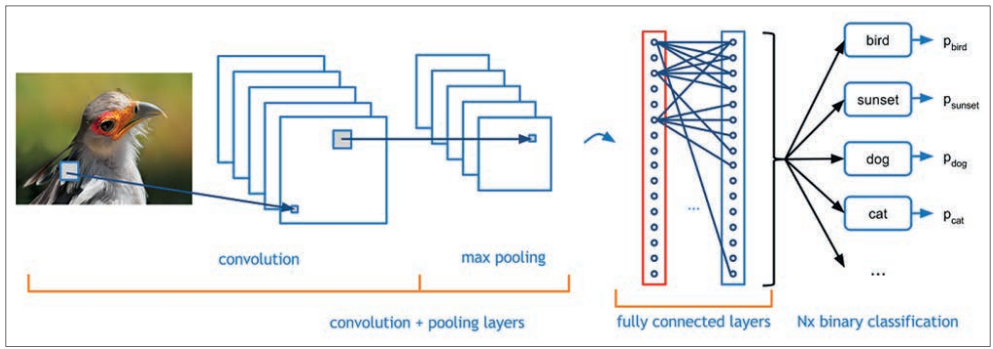

Source: Our processing on Deshpande, 2016

## 4. Background: Istat's Big Data Project on the ICT Survey

The annual "Survey on ICT Usage in Enterprises" ("ICT survey" for short) is carried out in Italy – as in many EU member states – under Eurostat regulations. It collects data on the usage of Information and Communication Technologies, the Internet, e-business and e-commerce in enterprises. The target population covers all the active enterprises with at least 10 employees. Enterprises with size 10-249 are sampled, whereas those with size 250 or more are all observed. The sampling design is one-stage stratified simple random sampling, with strata defined by crossing economic activity (NACE), enterprise size and geographical region (NUTS1). The sample is drawn from ASIA, the Istat archive of about 4.5 million Italian active enterprises. In the 2014 round of the survey, the planned sample size was about 30,000 units and the response rate was roughly 63%, yielding a respondent sample of nearly 19,000 enterprises (*i.e.* about 10% of the overall target population).

As discussed in Barcaroli *et al.* (2015 e 2016), Istat is actively investigating ways to:

1. Automatically scrape massive amounts of textual data from the websites of those enterprises, among the ones taking part to the ICT survey, which actually provided their website's URL through the survey questionnaire;

2. Train a machine learning algorithm to *learn* how to *predict* a survey variable $Y$ (*e.g.* whether the enterprise has e-commerce facilities deployed on its own website) using, as input information $X$, the text scraped from the enterprise website.

With respect to the second objective, only *supervised* learning approaches have been experimented so far. This means that the candidate machine learning algorithm is always fed with a *labelled training set* of ($Y_i$, $X_i$) pairs, where $Y_i$ is the *observed* survey value of the target variable of enterprise $i$, and $X_i$ is the corresponding web-scraped text. The performance of the trained algorithm is subsequently *tested* comparing its predictions $Y_j^*$ – based on a set of *unlabelled* $X_j$ values, which were never used before during training – to the corresponding *gold-standard* survey values $Y_j$.

Many learners have already been considered in Barcaroli *et al.* (2015 e 2016), ranging from statistical parametric models (the Logistic model), to ensemble learners (Random Forest, Adaptive Boosting, Bootstrap Aggregating), and including well known, traditional algorithms like Naïve Bayes and Support Vector Machines (SVM), as well as a new approach named SLAD (Statistical and Logical Analysis of Data). The obtained results for variable $Y =$ *"e-commerce (yes/no)"* are schematically reported in Table 2, along with a preliminary result of the new Deep Learning approach proposed here (highlighted in italic). Note that competing ML approaches have been ranked in Table 2 by *F-measure* (namely the harmonic mean of Precision and Recall), as the latter is the most reliable quality measure for the "e-commerce" classification task. This directly follows from the significant class imbalance of the gold-standard distribution of the target variable Y: 19% 'e-commerce' vs. 81% 'non-e-commerce'.

How to best encode the raw input data is a fundamental issue influencing the final performance of any ML algorithm. The optimal choice typically depends both on the learning task at hand and on specific characteristics qualifying the selected ML approach. In this respect, it must be stressed that all the learners studied in Barcaroli *et al.* (2015 e 2016) adopted the *same* encoding strategy, based on the traditional *bag-of-words* model. In such a model, which deliberately neglects word ordering and grammar rules, a text document is regarded as a simple set of terms and related frequencies. Therefore, a whole *corpus* of documents can be encoded into a Term-Document Matrix (TDM). Rows and columns of the TDM represent documents and terms occurring within them, respectively. The $(i, j)$ cell score of a TDM depends on the frequency of occurrence of term j within document i and (possibly) across the entire *corpus*. These scores can be computed according to multiple schemes, giving rise to TDMs of different flavors, *e.g. binary*, *frequency* and *Tf-Idf* (Term frequency-Inverse document frequency)3.

It is worth specifying that, despite the number of distinct terms occurring within the raw web-scraped text *X* was huge (several millions), the TDMs used in Barcaroli *et al.* (2015 e 2016) for the "e-commerce" classification task were always reduced to 1,000 columns, thanks to a preliminary heavy-filtering step. Besides standard information retrieval preprocessing (*e.g.* tokenisation, stop words removal, stemming and lemmatisation), this filtering step used Correspondence Analysis (see *e.g.* Benzécri, 1973) to hopefully identify the 1,000 words having higher predictive power on the values of the target variable *Y*.

---

3   The Tf-Idf scheme (Ramos, 2003) is often preferred in Text Mining applications, owing to its ability to increase the importance of a term proportionally to the term frequency inside the document, while penalising terms that are found to be very common in the whole *corpus*.

## 5. A new proposal based on Deep Learning

As anticipated in the Introduction, we propose to adopt cutting-edge Deep Learning techniques in order to address the prediction task of the ICT Big Data project. Our line of research, which is still ongoing, involves exploiting a CNN to solve the "e-commerce" classification problem. As this is essentially a supervised binary *text categorisation* problem, one could wonder why we decided to tackle it with a Deep Learning model originally designed for *image recognition*. The answer lies in the amazing *versatility* of DL models, which we mentioned in Section 3. As a matter of fact, there is a rich ongoing stream of scientific literature investigating the potentialities of CNNs in text classification (see *e.g.* Kim, 2014, and references therein).

Our work on CNN architectures encompassed two sequential phases. The first phase can be understood as a *feasibility study*, the second phase as the *actual implementation* of a production-ready processing pipeline. The next section explains the rationale of the feasibility study, describes its experimental setup and discusses its outcomes. The actual implementation of our proposal is illustrated in Section 7.

## 6. Feasibility study and preliminary results

Within the DL research community, CNNs are nowadays among the dominant approaches to text classification. Most of the recent literature in the field relies on *Word Embedding* models in order to encode input texts into a *rich* data representation that is actually able to capture many important *semantic* and *syntactic* relationships between words (see Section 7.1). Compelling evidence shows that word embedding models outperform more traditional text encoding techniques, like bag-of-words, in a wide variety of tasks. Moreover, the richer input data representation provided by these models turns out to be highly beneficial for CNNs.

Despite being aware of these findings, we decided *not* to rely on word-embedding techniques for our initial experiments. Instead, we committed ourselves to feed our CNN model with exactly the same input Term-Document Matrices used in Barcaroli *et al.* (2015 e 2016). This choice had a threefold objective: (*i*) it allowed us to directly compare our results to those already obtained with other ML approaches; (*ii*) it helped us gain familiarity with sophisticated DL techniques, by letting us initially focus on a low-complexity solution; and (*iii*) it freed us from the need of exploring different word-embedding setups and assessing their impact on the obtained results.

Going into technical details, the ICT Big Data project successfully scraped 10,164 enterprise websites, giving rise to TDMs of size (10,164 x 1,000). They were randomly split into a *training* TDM and a *testing* TDM of almost identical sizes: (5,082 x 1,000) and (5,083 x 1,000) respectively. This split was consistently preserved across experiments (*i.e.* different ML approaches) and TDM flavors (*i.e.* binary, frequency and Tf-Idf).

In order to feed our first CNN model, we *folded* each TDM's row (*i.e.* the bag-of-words content of the website of a given enterprise) into a square image of (32 x 32) = 1,024 pixels. Every pixel, within the image, "depicts" the TDM score of a given word, as illustrated in Figure 4.

**Figure 4 - Two images (32 x 32 = 1024 pixels) encoding alternative bag-of-words representations of the same web-scraped text (a). The left image is built upon one row of the *binary* TDM, the right one on the same row of the *Tf-Idf* version of the same TDM**

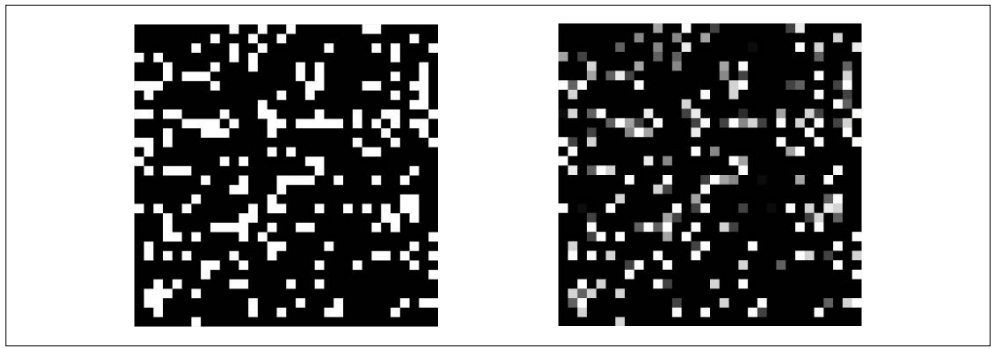

Source: Our processing
(a) The left image is built upon one row of the binary TDM, the right one on the same row of the Tf-Idf version of the same TDM.

For a *binary* TDM, a score of 1 (meaning *presence* of the word) is mapped to a *white* pixel, whereas a score of 0 (meaning *absence* of the word) is mapped to a *black* pixel. For *frequency* and *Tf-Idf* TDMs, the scores are mapped to *grayscale* colors (upon normalisation): the higher the score, the brighter the pixel tonality. Note that, since the size of our input images (*i.e.* 1,024 pixels) exceeded the number of columns (=words) of the TDMs that we inherited from Barcaroli *et al.* (2015 e 2016; *i.e.* 1,000 columns), we padded the residual 24 pixels with meaningless black pixels (*i.e.* artificial 0 scores).

Coming to the CNN model, for our feasibility study we chose to adopt a *LeNet* (LeCun *et al.*, 1989), which was appropriate to the moderate resolution of our input images. The architecture of this CNN is schematically illustrated in Figure 5. We encoded the results of the softmax output layer (Duan *et al.*, 2003) into a "one-hot" two component vector: <0,1> for 'e-commerce', and <1,0> for 'non-e-commerce'. Moreover, for our first bunch of experiments, we selected as loss function the Categorical Cross-Entropy (Botev *et al.*, 2007). As training algorithm, we focussed on Adam Gradient Descent (Adam) (Kingma and Ba, 2015). To build and evaluate our experimental CNN architecture we used the Keras Python Deep Learning API running on top of Theano (Chollet, 2015; Al-Rfou *et al.*, 2016).

**Figure 5 - The topology of the LeNet Convolutional Neural Network adopted in the feasibility study**

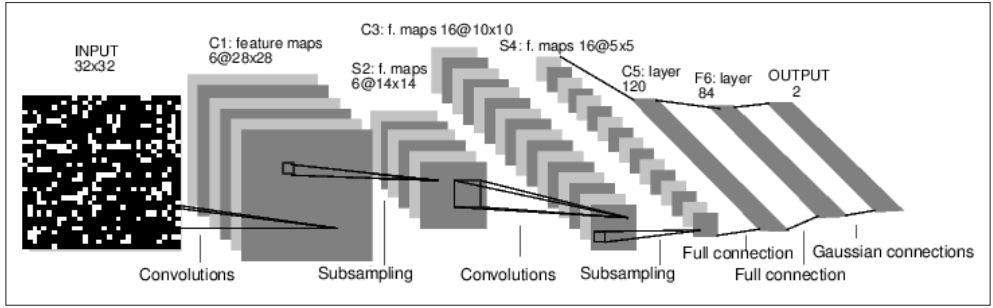

Source: Our processing on Deshpande, 2016

The top performance achieved by our first CNN model on the "e-commerce" classification problem is reported in Table 2, where it is highlighted in italic and compared to previous ML approaches (see Section 4). Note that our CNN attained its top performance when it was fed with *frequency* TDMs[4], consistently with previous findings concerning other ML models (Barcaroli *et al.*, 2016). The 0.57 F-measure score obtained by our LeNet architecture ranked our Deep Learning approach 3rd, only slightly below SVM (0.59) and SLAD (0.60). Recall that, owing to the substantial class imbalance affecting the target variable (19% 'e-commerce' vs. 81% 'non-e-commerce'), the *F-measure* is a much more reliable quality measure than *Accuracy* for the present task[5].

---

4   Binary and Tf-Idf TDMs led to slightly lower F-measure scores (~ 0.56), but we did not investigate these results in depth, because they were not essential to our feasibility study.

5   Moreover, the F-measure  $F = 2/(\mathrm{Prec}^{-1} + \mathrm{Rec}^{-1})$  is a *conservative* quality measure, as it can reach high values only when *both* Precision and Recall are high.

**Table 2 - Machine Learning approaches tested so far within Istat's Big Data project on the ICT survey (see also Section 4) (a)**

| ML approach | F-measure | Precision | Recall | Accuracy |
|---|---|---|---|---|
| SLAD | 0.60 | 0.58 | 0.62 | 0.84 |
| SVM | 0.59 | 0.55 | 0.64 | 0.83 |
| *Deep Learning* | *0.57* | *0.47* | *0.70* | *0.79* |
| Random Forest | 0.55 | 0.57 | 0.53 | 0.83 |
| Logistic | 0.53 | 0.53 | 0.53 | 0.83 |
| ANN | 0.52 | 0.52 | 0.52 | 0.82 |
| Boosting | 0.50 | 0.50 | 0.50 | 0.81 |
| Bagging | 0.48 | 0.53 | 0.44 | 0.82 |
| Naïve Bayes | 0.46 | 0.46 | 0.46 | 0.80 |

Source: Our processing
(a) The DL approach proposed in the feasibility study is highlighted in italic.

Although a 0.57 F-measure score may not seem an impressive result, we took it as a quite encouraging starting point for our Deep Learning line of research. The first reason was that, for the sake of comparability, we had until then restricted our CNN model to operate on bag-of-words text representations (*i.e.* Term-Document Matrices), which inevitably translated into cluttered images with very little spatial structure (as evident in Figure 4). The second reason was that the only ML approaches beating our initial proposal, *i.e.* SVM and SLAD, had required a fine-tuning of parameters to reach their top performances (as documented in Barcaroli *et al.*, 2016), something we had not yet done for our LeNet architecture.

Fortunately, we were confident that both issues affecting our initial CNN approach could be overcome. To reach this goal we decided, on the one hand, to dismiss the TDM approach and to switch to richer text representations, taking advantage of self-learned word-embeddings. Of course, we were aware that such a change would have required us to devise some smart *automatic summarisation* algorithm, in order to reduce the web-scraped text of each enterprise. Otherwise, the incorporation of word-embeddings would have inevitably led to very high resolution input images (see Section 7.1), with exploding computational costs. On the other hand, we planned to run more extensive experiments in order to optimise the hyperparameters of our candidate Deep Learning architectures, *e.g.* via grid search.

The next section shows how we turned the simple CNN architecture tested in the feasibility study into a much more sophisticated and better performing processing pipeline.

## 7. Processing pipeline

So far, we described our early attempts to solve the "e-commerce" classification task of the ICT Big Data project by means of Deep Learning techniques. Essentially, we fed a simple CNN architecture with web-scraped texts that previously underwent numerical encoding via the *bag-of-words* model. The overall approach was deliberately naïve, as appropriate for a feasibility study, but it yielded encouraging results. This section provides an overview of the work we carried out to design a more advanced DL pipeline that proved able to overcome the limits highlighted in Section 6.

Two main pillars underpin the processing pipeline we propose here. The first pillar is the adoption of *Word Embeddings* (WE). The second pillar is the adoption of a conceptual framework known as *False Positive Reduction* (FPR).

Next sections 7.1 and 7.2 introduce WE and FPR in isolation. A complete explanation of the way these pillars interact within our processing pipeline can be found in Section 7.3.

## 7.1 Word Embeddings

Modern WE models are generated by unsupervised learning algorithms – typically shallow neural networks, like Word2Vec (Mikolov *et al.*, 2013) or GloVe (Pennington *et al.*, 2014) – trained on very large text corpora. WE algorithms map words to vectors of a metric space in a very smart way, so that the resulting numeric representation of input texts effectively captures and preserves a wide range of semantic and syntactic relationships between words. Stated very simply: words that are strongly related from a syntactic and/or semantic point of view are mapped to embedding vectors that are almost parallel to each other; conversely, words that are syntactically and/or semantically loosely related are mapped to nearly perpendicular embedding vectors. Since the metric structure of embedding spaces is induced by the "cosine distance", WE algorithms are able to transform the notion of syntactic/semantic similarity between words into the notion of geometric closeness between the corresponding embedding vectors. This is an amazing achievement and clearly motivates the adoption of WE models as representational basis for many downstream Natural Language Processing tasks.

In principle, it is straightforward to use a WE model to turn a *text* into an *image*. Given a text of length $n$, that is an *ordered* sequence of words ($w_1$, …, $w_n$), and a WE space of dimension $d$, one simply encodes the text into a ($d$ x $n$) numeric matrix, whose $i$-th column represents the coordinates of the embedding vector of the $i$-th word $w_i$. This ($d$ x $n$) numeric matrix can, in turn, be seen as a grayscale image of ($d$ x $n$) pixels. In most applications, the dimension of the embedding space $d$ is set to values in the range 100-300 (Mikolov, *et al.*, 2013).

In practice, however, the very large size of our web-scraped input texts complicates the inclusion of a WE layer inside our processing pipeline. Indeed, each scraped websites amounts on average to $n \sim 10^4$ words and embedding vectors have usually $d \sim 10^2$ components, so that our input images would have typical size ($d$ x $n$)$\sim 10^6$, *i.e.* several megapixel. Because processing such high resolution images with a CNN would result in unaffordable computational costs, we had to develop a clever *automatic summarisation* algorithm to shorten our input texts. Technical details on this algorithm can be found in Section 7.3.1, where its role within the FPR approach will also become clear. For the time being, just note that the operation of *shortening* a text and then converting the obtained summary to an image via WE actually produces a *segmentation* of the image that would have been generated from the original text.

## 7.2 False Positive Reduction: Intuition and Rationale

The *False Positive Reduction* (FPR) conceptual framework is the second pillar underpinning our processing pipeline. Since – to the best of our knowledge – the FPR approach has seldom (if ever) been applied before to text classification tasks, we offer here an intuitive insight into its working mechanism. The rationale for the adoption of the FPR framework in our application scenario should also emerge from this preamble, whereas technical details are deferred to later sections.

FPR is a popular training modality in the field of biomedical image classification, *e.g.* computer detection of lung cancer from thorax CT scans (Ge *et al.*, 2005). Our interest in FPR lies in the fact that the task of training a ML algorithm to classify thorax CT scans as 'cancer' or 'non-cancer' actually shares many challenges with our objective of classifying websites as 'e-commerce' or 'non-e-commerce'.

*First*, thorax CT scans contain an enormous amount of data about complex anatomical structures (*e.g.* lungs, airways, vessels, soft tissues), whereas the relevant information for cancer detection is mostly concentrated within few very small image parts, namely pulmonary *nodules*[6]. In other words, from a ML perspective, CT scans have a very low *signal-to-noise ratio*. Web-scraped texts suffer exactly the same issue: they are huge collections of words (up to order 106), where sentences identifying a website as 'e-commerce' are invariably overwhelmed by background noise.

*Second*, CT scans are very high resolution images. Unless exceptional hardware resources are available, this makes almost impractical to analyse them *as a whole* with a CNN, due to exploding computational costs. As anticipated in Section 7.1, the same would happen to our huge web-scraped texts, if we tried to *directly* encode them into the richer data representation provided by modern WE models.

The FPR framework offers a viable solution to both the aforementioned challenges.

FPR encompasses three main steps: (1) *input data segmentation*, (2) *training on data-segments with inherited labels*, (3) *class prediction of original data*. Let us exploit again the problem of cancer detection from CT scans as a guide to highlight the purpose of each step.

1. *Input data segmentation*. A first algorithm identifies pulmonary *nodules* within thorax CT scans. This way, each input CT scan is mapped into *a set* of nodule images. Note that while *gold-standard* cancer/non-cancer labels coming from radiological diagnoses are available for thorax CT scans, *no* such labels are instead available for the derived nodules. Nevertheless, segmentation dramatically decreases the complexity of the original images, with a simultaneous striking gain in terms of signal-to-noise ratio.

2. *Training on data-segments with inherited labels*. After segmentation, a ML algorithm is trained to classify nodules (*instead* of thorax CT scans) as cancer/non-cancer. Since no knowledge is actually available about the benign or malignant nature of the nodules, each nodule of the training set can only *inherit* the label of the CT scan from which it originated. Therefore, *all* the nodules derived from a *positive* CT scan (*i.e.* a case of cancer) will be labelled as 'cancer', irrespective of their actual benign or malignant nature. As a result, many *negative* (*i.e.* benign) nodules of the training set will be *wrongly* annotated as *positives* (*i.e.* malignant) in the gold-standard. This proliferation of *False Positive* labels in the training set is the price one has to pay for the benefits of segmentation[7].

---

6   Note, indeed, that real-world radiological screening of lung cancer typically focusses on the analysis of pulmonary nodules.

3. *Class prediction of original data.* Once trained on nodules, the ML algorithm can natively only predict 'cancer' probabilities of *nodules*. This means that the ML predictions have to be suitably modified, if one wants to enable detection of lung cancer from thorax CT scans. A very simple, though often effective, adjustment amounts to assigning to each CT scan of the test set the highest 'cancer' probability observed among its nodules.

The key underlying assumption of the FPR framework is that the adopted ML algorithms are *tolerant* to misclassified training examples, *i.e.* they can achieve accurate predictions *despite* the proliferation of contaminating false positives induced by the segmentation step (whereby the name of the method: False Positive *Reduction*). In the biomedical field, this key assumption has been recently shown to hold for specific Deep Learning algorithms, *e.g.* the Multi-View ConvNets proposed in (Setio *et al.*, 2016).

## 7.3 Technical Implementation

Let us now switch to the technical implementation of our processing pipeline. The aim of this section is to describe: (*i*) how we took advantage of WE models, and (*ii*) how we deployed the FPR framework into our e-commerce detection application. To make the presentation easier, the implementation of each step of the FPR approach will be detailed in a dedicated subsection.

---

7 Note, incidentally, that *no False Negatives* can be generated as a byproduct of segmentation, as 'non-cancer' (*i.e.* negative) CT scans do not contain *any* malignant (*i.e.* positive) nodules.

### 7.3.1 Data Segmentation

In the *segmentation step* of the FPR approach, we map each scraped website into a small-cardinality set of *sentences* of equal length. These sentences are generated by an original automatic summarisation algorithm that we developed on purpose (as already mentioned in Sections 6 and 7.1). To identify relevant sentences within the text, the algorithm exploits a set of *marker words* with high discriminative power for the detection of e-commerce facilities in the website. Whenever one marker word is found in the text, all the surrounding words up to $k$ positions away from the marker are kept. This way, each website is segmented into a variable number of synthetic sentences of common length $2k + 1$. In our experiments, we typically use $k \sim 10$.

Evidently, the core of the summarisation algorithm lies in the way *marker words* are recruited. To perform this task, we implemented two methods of different complexity. For presentation convenience, in what follows we will call these methods the '*simple strategy*' and the '*advanced strategy*'. To automatically build the set of marker words that guide the segmentation, both strategies leverage *self-trained*[8] word-embeddings.

First, a WE algorithm is trained on the *corpus* of 5,082 web-scraped texts belonging to the training set of the ICT Big Data project. Then, a handful of e-commerce specific words (*e.g.* 'cart', 'account', 'pay', 'online') are selected as marker *seeds* by the user, and their embedding vectors are summed. The resulting sum vector is subsequently used as a "*bait*" to attract new words inside the markers set. The '*simple strategy*' and the 'advanced strategy' only differ in the way they use the "bait vector" to probe the embedding space and to attract new marker words. While the 'simple strategy' emphasises *exploitation*, the 'advanced strategy' emphasises *exploration*.

The '*simple strategy*' is straightforward: the $m$ embedding vectors that are *nearest* to the "bait vector" are identified in a single shot, and the corresponding words (*e.g.* 'order', 'wishlist', 'payment', 'checkout', 'shop', 'paypal', 'shipping', …) are directly recruited and join the ranks of the markers.

---

8   We also tested *pre-trained* word-embeddings (which we obtained with Word2Vec, using the Italian version of Wikipedia as training *corpus*) but they consistently *underperformed* self-trained ones.

The '*advanced strategy*', is more sophisticated. The underlying algorithm relies on graph theory and cannot be thoroughly explained in this paper (we documented it elsewhere; see De Fausti *et al.*, 2018).

What deserves to be stressed here is that the 'advanced strategy' does not identify the best *m* embedding vectors in a single shot. Marker words are instead recruited *progressively* during subsequent iterations. The goal is to perform a wider-range exploration of the neighborhood of the "bait vector". Of course, attention has been paid to prevent the algorithm from losing the initial semantic focus (set by the seed words) too quickly along the iterations.

Figure 6 provides one example of marker words generated by the '*simple strategy*' and by the '*advanced strategy*'. Note that, in both cases, the same 4 seed words ('online', 'carrello', 'ordina', 'commerce') have been used to define the "bait vector". The cardinality of the marker sets is also the same, *m* = 50. However, the results are noticeably different, as expected.

**Figure 6 - Two graphs depicting m = 50 marker words identified by the '*simple strategy*' (left graph) and by the '*advanced strategy*' (right graph) (a)**

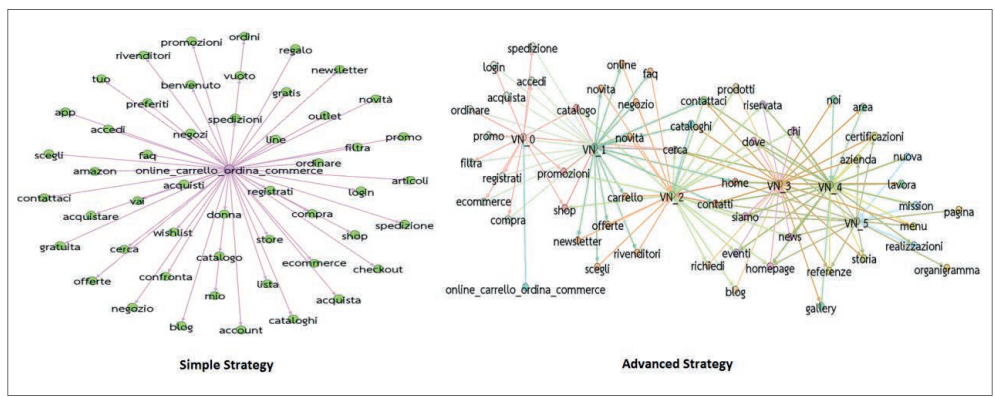

Source: Our processing
(a) Although the same 4 seed words ('online', 'carrello', 'ordina', 'commerce') have been used, the two strategies clearly led to different sets of markers words and, therefore, to different segmentations of scraped websites.

A few comments on the described segmentation strategies are in order. *First*, given the outstanding ability of word-embeddings to capture similarities between words, both strategies generate marker words characterised by good e-commerce *detection power*, likewise the original seeds. *Second*, marker words with high discriminative power for the detection of e-commerce help us *reduce*, since the beginning, the *False Positive* rate induced by the segmentation step of the FPR framework. That is, most of our segmented sentences will be *true* positives, akin to *malignant* nodules in the lung cancer analogy. *Third*, the algorithms that extract marker words are almost entirely *data-driven*: the analyst has only to provide few domain specific words as initial seeds. *Fourth*, the segmentation of websites into a set of synthetic sentences is entirely *automated*.

### 7.3.2 Training on Segments

After the segmentation step, our processing pipeline trains a CNN to classify *sentences* as 'e-commerce' or 'non-e-commerce' along the lines of Kim, 2014. Of course, as it is typical of the FPR approach, each sentence of the training set inherits its label $Y_i$ from the original website. Moreover, as sketched in Section 7.1, the text of each sentence is encoded into an ordered sequence of embedding vectors, namely a grayscale image $X_i$ of $(2k + 1)$ x $d$ pixels, being d the dimension of the embedding space. Note that, as $k \sim 10$ and $d \sim 10^2$, these images are of size $\sim 10^3$ and processing them with a CNN does not pose any computational problems.

The architecture of the CNN we designed is schematically illustrated in Figure 7. Observe that this is actually a specific kind of CNN, known as *Conv1D*. The choice of a Conv1D CNN is easily motivated as follows. Typically, the convolution layers of CNNs (see Figure 3 and Figure 5) involve learnable filters that slide over the *whole* input image by moving in 2D, *i.e.* through horizontal *and* vertical translations. Conv1D CNNs, instead, only allow the filters to slide in 1D, *e.g. just* vertically. Of course, in our application, filters must be constrained to only move *horizontally*: this way, they can only slide over *full* columns of the input image, namely over words (= embedding vectors). As a result, the integrity of embedding vectors is preserved. Note that this property is easily recognised in Figure 7, where it is conveyed by blue rectangles in the convolutional layers.

**Figure 7 - The topology of the Conv1D CNN used within the proposed False Positive Reduction framework (a)**

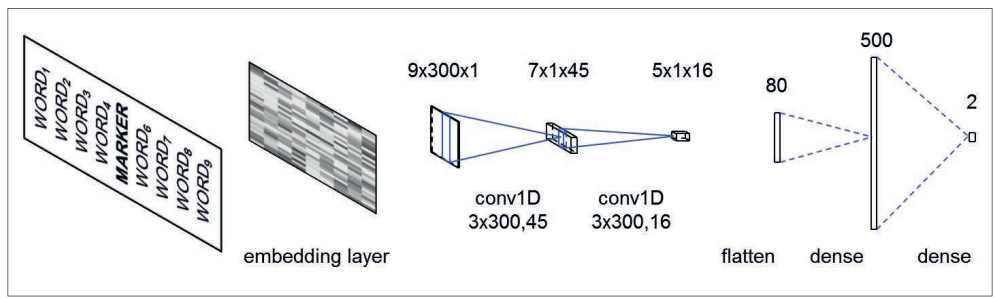

Source: Our processing
(a) In this figure $k$ = 4 is assumed, so that each website is segmented into a set of synthetic sentences of $(2k + 1)$ = 9 words. Moreover, the embedding vectors used to turn sentences into images have dimension d=300.

As shown in Figure 7, our Conv1D CNN has a receptive field of size 3 x 300, uses a stride of length 1, and involves two convolutional layers. The first convolutional layer learns 45 filters, the second one 16 filters, and no pooling layers have been inserted between them (unlike for the LeNet architecture tested in Section 6). The fully connected part of the network consists of two hidden layers. The number of neurons of the first fully connected layer is a function of the length of the input sentence $(2k + 1)$: in Figure 7, for instance, $k$=4 implies 80 neurons. The number of neurons of the second fully connected layer is instead fixed to 500. The output layer has two neurons, which return 'e-commerce' and 'non-e-commerce' probabilities of *sentences*.

We use ReLUs as activation functions for all the hidden layers of our Conv1D CNN model (be they convolutional or fully connected), and softmax for the output layer. As loss function, we adopt the Categorical Cross-Entropy. As training algorithm, we employ RMSProp with mini-batches of size 1,024 and a validation split of 25%.To prevent overfitting, training is performed for *very few* epochs, 10 at most, with an early stopping criterion ('patience') of 1 epoch. For the same reason, dropout regularisation is applied to all the layers with a *very high* rate of 50%. Indeed, while overfitting is a general concern in ML, it is even more serious in a FPR framework, because the training set is always contaminated by false positives introduced during the segmentation step. To implement and evaluate our model, we use Keras on top of Theano.

### 7.3.3 Class Prediction of Original Data

After having trained the Conv1D CNN on *sentences*, our processing pipeline suitably *modifies* its predictions, in order to enable the modified model to detect e-commerce at the *website* level. The goal is essentially to determine the 'e-commerce' probability of a website, given the predicted 'e-commerce' probabilities of its segments, *i.e.* sentences. We have explored and implemented two different methods to pursue this objective, which represents the final step of the FPR framework.

The *first method* is rather conventional: it simply assigns to each test website a predicted 'e-commerce' probability which equals the *highest* probability observed among its segmented sentences. For convenience, let us call this method the '*Max Rule*'.

The *second method* is original and more complex. It relies on a second dedicated Neural Network – specifically, a multilayer perceptron (MLP) – which is added to the pipeline just after the Conv1D CNN. This MLP is fed with labelled pairs $(Y_i, Z_i)$, where $Y_i$ is the 'e-commerce'/'non-e-commerce' status of the *i*-th website, and $Z_i$ is the *histogram* of the predicted probabilities of its segmented sentences. The MLP is then trained to find a hopefully *optimal decision rule* to map 'e-commerce' probabilities from the sentence level to the website level. The intuition behind the adoption of a dedicated MLP is, of course, that the optimal decision rule may actually turn out to be much more complicated than the naive '*Max Rule*', and could even depend on the *whole* distribution of sentence-wise probabilities of each website (which is summarised by the histogram). For the sake of conciseness, let us refer to this second method as the '*Histogram Rule*'.

The architecture of the MLP implementing the '*Histogram Rule*' is schematically illustrated in Figure 8. For consistency, the number of input neurons must of course match the number of bins used to construct the histograms. Thus, since we used 10 bins of equal width for the histograms, the input layer of the MLP contains 10 neurons. The MLP has 3 hidden layers, made up of 16, 8, and 4 neurons respectively. The output layer has two neurons, which return 'e-commerce' and 'non-e-commerce' probabilities of *websites*.

**Figure 8 - The topology of the MLP used to implement the 'Histogram Rule' (a)**

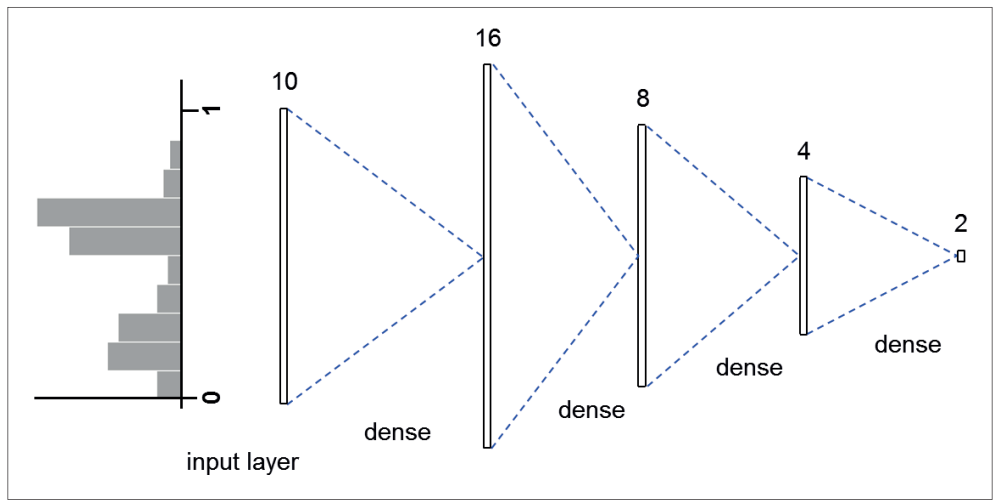

Source: Our processing
(a) The MLP learns to reconstruct the 'e-commerce' probability of a website, given the predicted 'e-commerce' probabilities of its constituting segments, *i.e.* sentences. The sentence-level probabilities are fed to the MLP through a histogram with 10 equal-width bins.

All the hidden layers of the MLP use ReLUs as activation functions, while the output layer uses a softmax. The Categorical Cross-Entropy is employed as loss function, and RMSProp as optimisation algorithm, with mini-batches of size 600 and a validation split of 40%. Training is performed for at most 800 epochs, with an early stopping criterion ('patience') of 1 epoch. To prevent overfitting, L2 regularisation (with lambda $8 \cdot 10\text{-}6$) and dropout (with rate 25%) have been applied. To implement and evaluate our model, we use Keras on top of Theano. Observe that overfitting (albeit still undesirable) is not as dire a threat for the MLP as it was for the Conv1D CNN. This is because the Conv1D CNN was trained on *sentences* whose "e-commerce" labels were *inherited* and therefore *contaminated* by false positives, whereas the MLP is trained on *genuine* and *uncontaminated* "e-commerce" labels of *websites*. This explains why more epochs and lower dropout rates are used for the MLP as compared to the Conv1D CNN.

Both the '*Max Rule*' and the '*Histogram Rule*' eventually return predicted 'e-commerce' *probabilities* of websites. But the final output of our processing pipeline must actually be the *decision* of classifying each test websites as either 'e-commerce' or 'non-e-commerce'. To this end, a *classification threshold* has to be set, which allows for an additional degree of freedom within our pipeline. The simplest choice would be to set the threshold to 0.5: this would assign each website to the class to which it has the highest predicted probability of belonging. A more sophisticated alternative could be to explicitly take into account the inherent *class imbalance* of the "e-commerce" distribution, as known by the *training set* (about 19% 'yes' vs. 81% 'no'). In this case, the classification threshold could be *adjusted* to best reproduce the gold-standard proportion of 'e-commerce' observed in the *training set*. Both these classification thresholds have been implemented in our processing pipeline: in what follows, we will refer to them as the '*Unadjusted Threshold*' and the '*Adjusted Threshold*' respectively. Note that the '*Adjusted Threshold*' method has an interesting side-effect: it tends to equalize the Precision and the Recall of the classifier, which often results in a better F-measure score than the one induced by the '*Unadjusted Threshold*'.

## 8. Experiments

In this section, we empirically evaluate the performance of the Deep Learning processing pipeline documented in Section 7. We also provide insights on the impact of the most important tunable hyperparameters of our approach.

Likewise the feasibility study of Section 6, experiments focus here on the "e-commerce" prediction task of Istat's ICT Big Data project. Recall that available data amount to 10,164 pairs $(Y_i, X_i)$, where $Y_i$ is the "e-commerce" label of enterprise $i$ (derived from the traditional ICT survey), and $X_i$ is the corresponding web-scraped text. Recall also that these 10,164 pairs were randomly split into a *training set* and a *test set* of almost identical sizes: 5,082 and 5,083 respectively.

Word Embeddings (WE) are the first pillar underpinning our proposal: they provide the representational basis to turn texts into images (Section 7.1), and play a central role in the automatic summarisation algorithm that segments scraped websites into synthetic sentences (Section 7.3.1). Our pipeline only relies on *self-trained* WE, which we obtained with Word2Vec, using as training *corpus* the collection of 5,082 web-scraped texts of the *training set*. Note that our approach conventionally maps to the *null vector* of the embedding space those words belonging to websites of the *test set* that happen to be *absent* from the vocabulary of self-learned WE. We report here the configuration of **Word2Vec**'s main parameters (Levy *et al.*, 2015) that we settled in our experiments:

- Embedding space dimension: **d = 300**.

- Window size: **8 words**.

- Learning model: **CBOW with negative sampling**.

Because our pipeline is quite sophisticated and involves several hyperparameters that can be tuned to improve classification performance, we carried out an extensive grid-search. To make the presentation of the grid-search results easier, we provide here a synopsis of the hyperparameters we tested (their meaning is documented in Sections 7.3.1, 7.3.2 and 7.3.3). The synopsis below lists the hyperparameter values that have been explored (reported in bold), along with the logical building-blocks to which each hyperparameter belongs (reported in italic):

1.  *Marker words selection*:                  'Simple Strategy' / 'Advanced Strategy'

2.  *Marker words cardinality*:              m = (50, 100, 150)

3.  *Sentence width (2k + 1)*:                k = (2, 4, 8)

4.  *Website probability reconstruction*: 'Max Rule' / 'Histogram Rule'

5.  *Website classification threshold*:      'Unadjusted' / 'Adjusted'

Hyperparameters 1 and 2 govern the way marker words are automatically recruited starting from few seed words specified by the user. In all experiments we used the *same 4 seed words* (see Figure 6): **'online'**, **'carrello'**, **'ordina'**, **'commerce'**.

According to the synopsis above, our grid-search probed (2*3*3*2*2) = 72 distinct hyperparameter configurations. For each configuration, we tested the performance of our processing pipeline on the ICT Big Data project. Even after setting all the hyperparameters to specific values, the results of a Deep Learning algorithm are always affected by some residual random variability. This is because *e.g.* the network weights are initialised randomly, and stochastic optimisation algorithms are used to optimise the weights during training. To control for the impact of this residual variability on the performance of our algorithms, we performed 50 runs of our processing pipeline for each hyperparameter configuration tested in the grid-search. Therefore, the grid-search eventually resulted in (72*50) = 3,600 runs: for each run we measured the quality of obtained results in terms of Precision, Recall, F-measure and Accuracy. Note that, for each run, we *re-trained* both the Conv1D CNN and (if required by the hyperparameter configuration being tested) the MLP.

**Table 3 - F-measure scores obtained in the grid-search, conditional to tested hyperparameter values (a)**

| Hyperparameter | Value | F-measure Distribution Summary | | | | | | Nobs |
|---|---|---|---|---|---|---|---|---|
| | | Min | 1st Q | Median | Mean | 3rd Q | Max | |
| Marker words selection | Simple Strategy | 0.02 | 0.60 | 0.62 | 0.60 | 0.67 | 0.71 | 1,800 |
| | Advanced Strategy | 0.33 | 0.61 | 0.65 | 0.61 | 0.68 | 0.73 | 1,800 |
| Marker words cardinality | m = 50 | 0.04 | 0.59 | 0.62 | 0.60 | 0.67 | 0.71 | 1,200 |
| | m = 100 | 0.02 | 0.61 | 0.63 | 0.61 | 0.67 | 0.71 | 1,200 |
| | m = 150 | 0.04 | 0.61 | 0.65 | 0.61 | 0.68 | 0.73 | 1,200 |
| Sentence width (2k + 1) | k = 2 | 0.02 | 0.60 | 0.62 | 0.60 | 0.67 | 0.71 | 1,200 |
| | k = 4 | 0.04 | 0.61 | 0.64 | 0.61 | 0.68 | 0.73 | 1,200 |
| | k = 8 | 0.04 | 0.61 | 0.63 | 0.61 | 0.67 | 0.71 | 1,200 |
| Website probability reconstruction | Max Rule | 0.44 | 0.53 | 0.60 | 0.59 | 0.63 | 0.69 | 1,800 |
| | Histogram Rule | 0.02 | 0.64 | 0.67 | 0.65 | 0.68 | 0.73 | 1,800 |
| Website classification threshold | Unadjusted | 0.44 | 0.44 | 0.62 | 0.56 | 0.65 | 0.73 | 1,800 |
| | Adjusted | 0.02 | 0.61 | 0.66 | 0.65 | 0.68 | 0.72 | 1,800 |
| – | – | *0.02* | *0.61* | *0.63* | *0.61* | *0.68* | *0.73* | *3,600* |

Source: Our processing
(a) For convenience, the overall (*i.e.* unconditional) F-measure distribution is reported in italic in the last row. Column 'Nobs' gives the number of grid-search runs performed while holding fixed a given hyperparameter.

Table 3 above condenses the F-measure scores we obtained from the grid-search. More precisely, the table reports a summary of the (*conditional*) F-measure distribution obtained for each tested hyperparameter value. The overall (*i.e. unconditional*) F-measure distribution is also reported for convenience (last row, in italic).

With a top F-measure score of 0.73 reached in the grid-search, our pipeline clearly outperforms previous ML approaches listed in Table 2 of Section 6, and even subsequent attempts reported in Barcaroli and Scannapieco 2019 (top F-measure 0.63, achieved using Random Forest on comparable – albeit not identical – data collected in the 2017 round of the ICT survey). Note that this is a very robust result, as the top performances of competing ML approaches consistently lie below the average (and median) performance level attained by our proposal (despite occasional "unfortunate" runs occurred in the grid-search).

The same information provided in Table 3 is conveyed visually in Figure 9, which shows boxplots of F-measure scores by hyperparameter values.

Different values of the same hyperparameter correspond to boxplots of the same color. The *marginal effect* of each tested hyperparameters on the F-measure scores obtained in the grid-search emerges quite clearly from the plot.

**Figure 9 - Boxplots of F-measure scores by hyperparameter values. Boxplots related to different values of the same hyperparameter share the same color (a)**

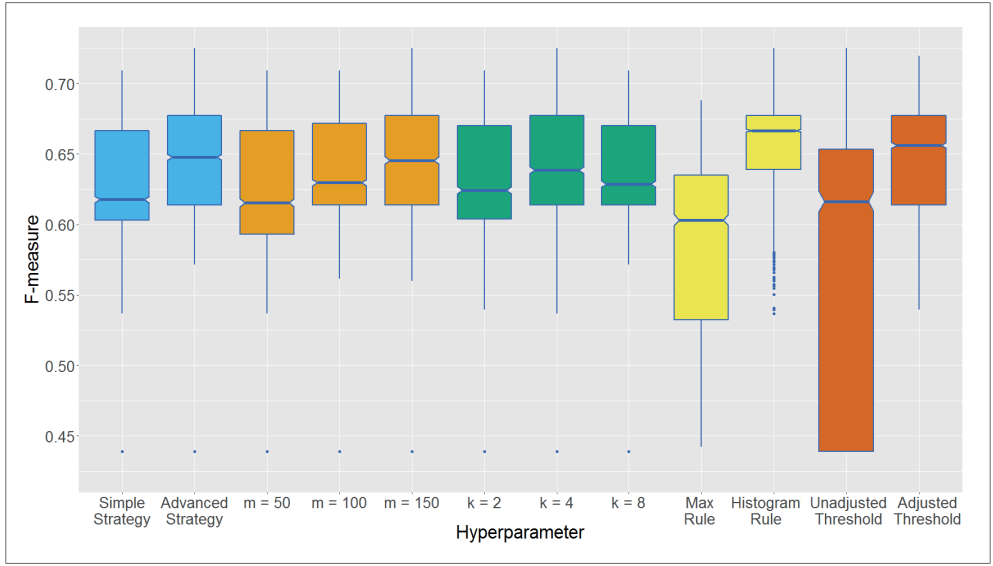

Source: Our processing
(a) Boxplots related to different values of the same hyperparameter share the same color. This plot illustrates the marginal effect of each tested hyperparameters on the F-measure scores obtained in the grid-search.

The evidence from Figure 9 can be summarised as follows:

1. The 'Histogram Rule' largely outperforms the naïve 'Max Rule'. This fully repays the design and computational extra costs of including an additional MLP in our pipeline.

2. The 'Adjusted Threshold' stands out as a much better alternative than the 'Unadjusted Threshold'. This comes as no surprise, given the substantial class imbalance of the "e-commerce" distribution.

3. The 'Advanced Strategy' for recruiting marker words that guide the segmentation is significantly better than the 'Simple Strategy'.

4. Performance monotonically grows with the number of marker words, at least in the range explored in the grid-search, m = (50, 100, 150). This suggests that a further increase of *m* might lead our pipeline to even higher F-measure scores. However, a growth in *m* results in heightened computational costs. Therefore, more experiments would be needed to find an optimal tradeoff limit for *m*.

5. Sentences of intermediate length (9 words, k = 4) perform better than either shorter (5 words, k = 2) and longer (17 words, k = 8) ones.

Up to now we only analysed the *marginal effects* of each tested hyperparameters on the F-measure scores obtained in the grid-search. However, the main reason to perform a grid-search is to evaluate the *interaction effects* of all hyperparameters, namely to find the best possible *combination* of hyperparameter values. Due to space restrictions, we cannot report here performance statistics for all the 72 configurations explored in the grid-search. Instead, we study in some detail the behavior of the top-2 configurations. The hyperparameter configurations that achieved highest maximum F-measure in the grid-search are the following:

- **TOP-1** (maximum F-measure = 0.73):
  Advanced Strategy | m = 150 | k = 4 | Histogram Rule | Unadjusted Threshold.

- **TOP-2** (maximum F-measure = 0.72):
  Advanced Strategy | m = 150 | k = 4 | Histogram Rule | Adjusted Threshold.

The TOP-1 and TOP-2 configurations only differ in the way they set the final website classification threshold. Somewhat unexpectedly in the light of observation (*ii*) above, TOP-1 does *not* adjust the threshold. However, one should not overlook that TOP-1 and TOP-2 are actually very close in both maximum F-measure and maximum Accuracy. This can be seen in Table 4.

**Table 4 - F-measure and Accuracy scores of the top-2 hyperparameter configurations found in the grid-search. For both configurations 50 runs were executed**

| Hyperparameter Configuration | Performance Measure | Distribution Summary | | | | | |
|---|---|---|---|---|---|---|---|
| | | Min | 1st Q | Median | Mean | 3rd Q | Max |
| TOP-1 | F-measure | 0.67 | 0.68 | 0.69 | 0.69 | 0.70 | 0.73 |
| TOP-2 | F-measure | 0.67 | 0.69 | 0.70 | 0.70 | 0.71 | 0.72 |
| TOP-1 | Accuracy | 0.86 | 0.88 | 0.88 | 0.88 | 0.89 | 0.90 |
| TOP-2 | Accuracy | 0.87 | 0.88 | 0.89 | 0.88 | 0.89 | 0.89 |

Source: Our processing

Table 4 shows subtle but interesting differences between the F-measure and Accuracy distributions of TOP-1 and TOP-2. These differences can be more easily recognised visually through Figure 10, where TOP-1 and TOP-2 are compared using F-measure and Accuracy density plots. There, blue continuous lines identify TOP-1 density plots and red dashed lines identify TOP-2 ones.

**Figure 10 - Density plots of F-measure (left panel) and Accuracy (right panel) scores for the top-2 hyperparameter configurations found in the grid-search (a)**

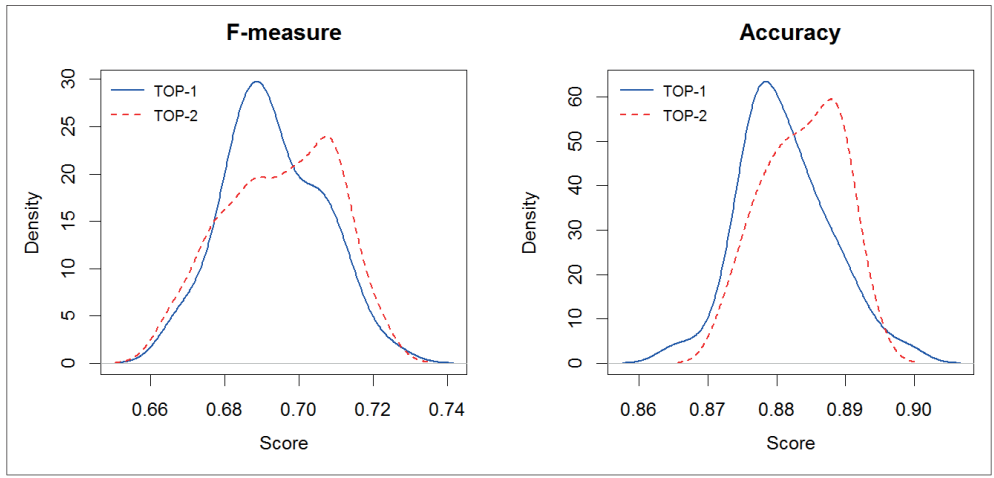

Source: Our processing
(a) The TOP-1 configuration densities are plotted with a blue continuous line; the TOP-2 ones, with a red dashed line. For both configurations 50 runs were executed.

The F-measure distribution of TOP-2 is clearly more concentrated at higher scores than the one of TOP-1, and has equal variance. The Accuracy distributions exhibit a similar pattern: TOP-2 peaks at higher scores, but is also less dispersed than TOP-1. Overall, the classification performance of TOP-2 appears at least as good as the one of TOP-1, but more stable. In the light of these considerations, and taking into account that the maximum F-measure and Accuracy scores of TOP-2 are only slightly below those of TOP-1, we are led to the conclusion that the best configuration of our processing pipeline is actually TOP-2.

Coming back to Table 4, it is worth stressing that all the *top* performances of previous ML approaches (listed in Table 2 of Section 6) consistently lie *below the minimum* performance level attained in 50 runs by the best configuration of our pipeline (*i.e.* TOP-2). This is a clear indication of the scale of the improvement achieved by our Deep Learning approach. We conclude the section by reporting below the top Precision, Recall, F-measure and Accuracy scores reached by our pipeline in the grid-search.

**Table 5 - Best hyperparameter configuration and top performance of the Deep Learning processing pipeline proposed in the work**

| Best Hyperparameter Configuration | F-measure | Precision | Recall | Accuracy |
|---|---|---|---|---|
| • Advanced Strategy<br>• m = 150<br>• k = 4<br>• Histogram Rule<br>• Adjusted Threshold | 0.72 | 0.73 | 0.72 | 0.89 |

Source: Our processing

As shown by Table 5 and Table 2, our processing pipeline achieved a relative gain in classification performance of +26% in terms of F-measure and +13% in terms of Accuracy with respect to our naïve model of Section 6.

With respect to the best competitor among previously tested ML approaches, the relative classification gain set by our proposal is +20% in terms of F-measure and +6% in terms of Accuracy.

Finally, it is worthwhile to note that our proposal even outperforms subsequent results achieved by Barcaroli and Scannapieco 2019 using Random Forest on comparable, albeit not identical, data collected in the 2017 round of the ICT survey. These authors report a top F-measure of 0.63 and a top Accuracy of 0.83. Therefore, with respect to these scores, our pipeline still seems ahead, with  relative gains of +14% in terms of F-measure and +7% in terms of Accuracy.

## 9. Ongoing work and conclusions

In this paper, we showed that Deep Learning techniques can successfully address the "e-commerce" prediction task of Istat's ICT Big Data project. To reach this goal, we developed a sophisticated processing pipeline and evaluated its performance through extensive experiments. Empirical evidence shows that our proposal outperforms all the alternative Machine Learning solutions already tested in Istat for the same task. Besides good classification performance, our pipeline exhibits other desirable properties:

1. *It is entirely automated*. Useful text features for the detection of e-commerce are learned by the CNN without any human intervention, directly from the Word Embedding representation of scraped websites.

2. *It is almost entirely data-driven*. The only domain knowledge assumed is required to identify a handful of e-commerce specific words to be used as initial seeds by the automatic summarisation algorithm.

3. *It is generalizable*. The applicability to other binary classification tasks, besides "e-commerce", is obvious. Only minor adjustments would be required to enable multinomial classification of websites (essentially, the user would have to pass to the system class-specific seed words).

4. *Can take advantage of non-textual input*. Since CNNs can process both texts and images, our pipeline can be extended to leverage also the images scraped from the websites of Italian enterprises.

At the moment, our research focusses on point (4.). The rationale behind this line of research is that an ensemble of two Deep Learning classifiers – the first one extracting features from texts, the second from pictures – could achieve better predictive accuracy than either of the original classifiers. Let us hint at how our FPR approach may be applied to digital images scraped from enterprise websites. Since a large part of the paper has already been devoted to the FPR framework, we focus here only on the *differences* between processing texts and processing digital images.

*Segmentation*. As digital images embedded into most websites are easily identified within HTML files through their filename extensions (*e.g.* .jpeg, .gif, .bmp, etc.), segmentation of websites into images is straightforward and can be directly performed by the web-scraping system.

*Image Scaling*. Web-scraped images come up in widely varying sizes. This is at odds with the constant length synthetic sentences we extracted from scraped texts. Since CNNs generally benefit from constant size input examples, all the scraped images have to be scaled (up or down) to a common format, *e.g.* to square images of size 256 x 256 pixels.

*DL architecture*. We are currently studying and testing the *Residual Neural Networks* (ResNet) models proposed in (He *et al.*, 2016), which are our best candidate until now. ResNet are very deep and sophisticated CNN architectures that achieved record-breaking accuracies in Computer Vision, even outperforming humans in some image recognition tasks.

*Training modality*. Given the wide variety of goods and services that Italian enterprises can either sell (in the 'e-commerce' case) or only showcase (in the 'non-e-commerce' case) on their websites, our web-scraped images will span a tremendous range of diverse subjects. As a consequence, it will be extremely hard to train a CNN to either (*i*) identify few special images having high 'e-commerce' detection power (*e.g.* shopping cart icons), or (*ii*) discover useful latent correlations between the e-commerce/non-e-commerce status of websites and image subjects (*e.g.* smartphones are more frequently sold online than pets). ResNet models have the potential to cope with such complex problems, but they need *enormous* training sets of labelled examples to reach successful results (*e.g.* the ImageNet database[9] stores *over ten million* hand-annotated images). This is a serious concern, since our ICT training set only amounts to *few thousands* labelled websites. We believe that *Transfer Learning* (see *e.g.* Pan and Yang, 2010) can be a viable countermeasure to this issue. The base idea of Transfer Learning is to first train a CNN on a different (though not entirely unrelated) problem for which a huge training set is available, and to subsequently exploit the pre-trained model as a starting point to solve the real problem of interest, for which limited labelled examples are available. The intuition is that the knowledge gained in the first phase (and stored inside the weights of the pre-trained network) could be further enriched and fine-tuned in the second phase, therefore boosting the final prediction accuracy of the CNN. At the moment, we are running experiments in which a ResNet model pre-trained on ImageNet is retrained to classify our web-scraped images as 'e-commerce'/'non-e-commerce'.

---

9   ImageNet website: http://www.image-net.org/.